\documentclass[10pt,twocolumn,letterpaper]{article}

\usepackage{cvpr}              

\DeclareMathAlphabet\mathbfcal{OMS}{cmsy}{b}{n}

\def\0{{\bf 0}}
\def\1{{\bf 1}}

\def\bT{{\bf T}}

\def\bp{{\bf p}}

\def\bz{{\bf z}}

\def\bp{{\bf p}}

\def\bz{{\bf z}}

\def\eg{\emph{e.g.}} 
\def\ie{\emph{i.e.}} 
 
 \def\vs{\emph{vs.}}

\usepackage{amsmath}

\usepackage{graphicx}
\usepackage{amsmath}
\usepackage{amssymb}
\usepackage{booktabs}
\usepackage{booktabs}
\usepackage{xcolor}
\usepackage{colortbl}
\usepackage{bbding}
\usepackage{multirow}
\usepackage{caption}
\usepackage[accsupp]{axessibility}
\definecolor{mygray}{rgb}{0.9,0.9,0.9}
\definecolor{darkgreen}{rgb}{0.1, 0.7, 0.1}
\definecolor{blackpink}{rgb}{0.83, 0.19, 0.79}
\definecolor{orange}{rgb}{0.87, 0.3, 0.14}

\def\cph{\textcolor{black}}
\def\reb{\textcolor{black}} 
\def\rebb{\textcolor{black}} 
\def\mk{\textcolor{black}}

\def\todo{\textcolor{black}}

\usepackage{paralist}

\usepackage[pagebackref,breaklinks,colorlinks]{hyperref}

\usepackage[capitalize]{cleveref}
\crefname{section}{Sec.}{Secs.}
\Crefname{section}{Section}{Sections}
\Crefname{table}{Table}{Tables}
\crefname{table}{Tab.}{Tabs.}
\newcommand{\fullname}{Masked Motion Encoding\xspace}
\newcommand{\fullnamelite}{masked motion encoding\xspace}
\newcommand{\sexyname}{MME\xspace}
\newcommand{\para}{\vspace{0.25cm}}

\newcommand{\key}{important temporal clues\xspace}

\def\cvprPaperID{1550} 
\def\confName{CVPR}
\def\confYear{2023}

\begin{document}

\title{\fullname for Self-Supervised Video Representation Learning}

\author{
    Xinyu Sun\textsuperscript{\rm 1 \rm 2}\thanks{Equal contribution. Email: \{csxinyusun, phchencs\}@gmail.com} ~~ 
    Peihao Chen\textsuperscript{\rm 1}\footnotemark[1] ~~ 
    Liangwei Chen\textsuperscript{\rm 1} ~ 
    Changhao Li\textsuperscript{\rm 1} \\
    Thomas H. Li\textsuperscript{\rm 6} ~
    Mingkui Tan\textsuperscript{\rm 1 \rm 5}\thanks{Corresponding author. Email: mingkuitan@scut.edu.cn} ~
    Chuang Gan\textsuperscript{\rm 3 \rm 4} \\
    \textsuperscript{\scriptsize{\rm 1}}\small{South China University of Technology,}
    \textsuperscript{\rm 2}\small{Information Technology R\&D Innovation Center of Peking University} \\
    \textsuperscript{\scriptsize{\rm 3}}\small{UMass Amherst,}
    \textsuperscript{\rm 4}\small{MIT-IBM Watson AI Lab,}
    \textsuperscript{\rm 5}\small{Key Laboratory of Big Data and Intelligent Robot, Ministry of Education} \\
    \textsuperscript{\scriptsize{\rm 6}}\small{Peking University Shenzhen Graduate School} \\
}

\maketitle

\begin{abstract}
How to learn discriminative video representation from unlabeled videos is challenging but crucial for video analysis. The latest attempts seek to learn a representation model by predicting the appearance contents in the masked regions. However, simply masking and recovering appearance contents may not be sufficient to model temporal clues as the appearance contents can be easily reconstructed from a single frame. To overcome this limitation, we present \fullname (\sexyname), a new pre-training paradigm that reconstructs both appearance and motion information to explore temporal clues. In \sexyname, we focus on addressing two critical challenges to improve the representation performance: 1) how to well represent the possible long-term motion across multiple frames; and 2) how to obtain fine-grained temporal clues from sparsely sampled videos. Motivated by \todo{the fact that human is able to recognize an action by tracking objects' position changes and shape changes}, we propose to reconstruct a motion trajectory that represents these two kinds of change in the masked regions. Besides, given the sparse video input, we enforce the model to reconstruct dense motion trajectories in both spatial and temporal dimensions. Pre-trained with our \sexyname paradigm, the model is able to anticipate long-term and fine-grained motion details. Code is available at \url{https://github.com/XinyuSun/MME}.

\end{abstract}


\section{Introduction}
\label{sec:intro}
Video representation learning plays a critical role in video analysis like action recognition~\cite{MTV,Gan_2020_ECCV,chen2020regnet}, action localization~\cite{RAM,zeng2019breaking}, video retrieval~\cite{MSR-VTT,Zeng_2020_CVPR}, videoQA~\cite{huang2020LGCN}, \emph{etc}. 
Learning video representation is very difficult for two reasons. Firstly, it is extremely difficult and labor-intensive to annotate videos, and thus relying on annotated data to learn video representations is not scalable. Also, the complex spatial-temporal contents with a large data volume are difficult to be represented simultaneously. 
How to perform self-supervised videos representation learning only using unlabeled videos has been a prominent research topic~\cite{RSPNet,SpeedNet,VCP}.

Taking advantage of spatial-temporal modeling using a flexible attention mechanism, vision transformers~\cite{vit, TimeSformer, ViViT, Motionformer, MViT} have shown their superiority in representing video. 
Prior works~\cite{BEiT,MAE,iBOT} have successfully introduced the mask-and-predict scheme in NLP~\cite{GPT3,BERT} to pre-train an image transformer.
These methods vary in different reconstruction objectives, including raw RGB pixels~\cite{MAE}, hand-crafted local patterns~\cite{MaskFeat}, and VQ-VAE embedding~\cite{BEiT}, all above are static appearance information in images.
Based on previous successes, some researchers~\cite{BEVT,MaskFeat,VideoMAE} attempt to extend this scheme to the video domain, where they mask 3D video regions and reconstruct appearance information. 
However, these methods suffers from two limitations. \reb{\textbf{First}, as the appearance information can be well reconstructed in a single image with an extremely high masking ratio (85\% in MAE~\cite{MAE}), \todo{it is also feasible to be reconstructed in the tube-masked video frame-by-frame and neglect to learn \key.} This can be proved by our ablation study (cf. Section~\ref{exp:revisit}).}
\textbf{Second}, existing works~\cite{VideoMAE,MaskFeat} often sample frames sparsely with a fixed stride, and then mask some regions in these sampled frames. The reconstruction objectives only contain information in the sparsely sampled frames, and thus are hard to provide supervision signals for learning fine-grained motion details, \reb{which is critical to distinguish different actions~\cite{TimeSformer,ViViT}.}

\begin{figure*}[t]
	\centering
     \includegraphics[width=0.95\textwidth]{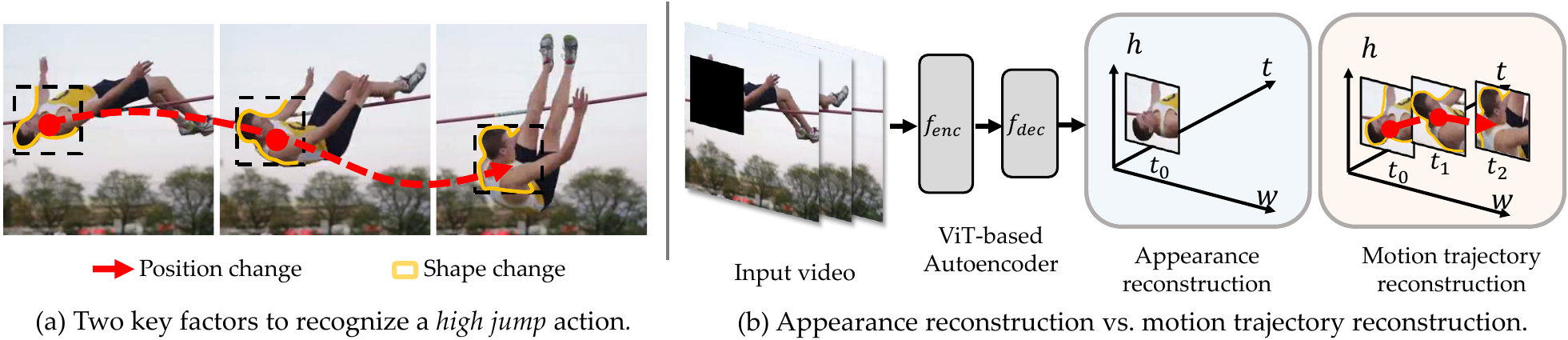}
     \caption{\textbf{Illustration of motion trajectory reconstruction for \fullname.} (a) Position change and shape change over time are two key factors to recognize an action, we leverage them to represent the motion trajectory. (b) Compared with the current appearance reconstruction task, our motion trajectory reconstruction takes into account both appearance and motion information.}
     \label{fig:ssv2-vis}
    \vspace{-2mm}
\end{figure*}

In this paper, we aim to design a new mask-and-predict paradigm to tackle these two issues. Fig.~\ref{fig:ssv2-vis}(a) shows two key factors to model an action, \ie, position change and shape change. \rebb{By observing the position change of the person, we realize he is jumping in the air, and by observing the shape change that his head falls back and then tucks to his chest, we are aware that he is adjusting his posture to cross the bar. We believe that anticipating these changes helps the model better understand an action.}

Based on this observation, instead of predicting the appearance contents, we propose to predict \emph{motion trajectory}, which represents \textbf{impending} position and shape changes, for the mask-and-predict task. \reb{Specifically, we use a dense grid to sample points as different object parts, and then track these points using optical flow in adjacent frames to generate trajectories, as shown in Fig.~\ref{fig:ssv2-vis}(b).} The motion trajectory contains information in two aspects: the position features that describe relative movement; and the shape features that describe shape changes of the tracked object along the trajectory. \reb{To predict this motion trajectory, the model has to learn to reason the semantics of masked objects based on the visible patches, and then learn the correlation of objects among different frames and try to estimate their accurate motions.} We name the proposed mask-and-predict task as \textbf{\fullname} (\sexyname). 

Moreover, to help the model learn fine-grained motion details, we further propose to interpolate the motion trajectory. Taking sparsely sampled video as input, the model is asked to reconstruct spatially and temporally dense motion trajectories. This is inspired by the video frame interpolation task~\cite{Zooming} where a deep model can reconstruct dense video at the pixel level from sparse video input. 
Different from it, we aim to reconstruct the fine-grained motion details of moving objects, which has higher-level motion information and is helpful for understanding actions. Our main contributions are as follows:

\begin{itemize}
    \item \mk{\todo{Existing mask-and-predict task based on appearance reconstruction is hard to learn \key}, which is critical for representing video content. 
    Our \fullname (\sexyname) paradigm overcomes this limitation by asking the model to reconstruct motion trajectory. 
    }

    \item \reb{Our motion interpolation scheme takes a sparsely sampled video as input and then predicts dense motion trajectory in both spatial and temporal dimensions. This scheme endows the model to capture long-term and fine-grained motion clues from sparse video input.
    }

\end{itemize}
Extensive experimental results on multiple standard video recognition benchmarks prove that the representations learned from the proposed mask-and-predict task achieve state-of-the-art performance on downstream action recognition tasks. 
Specifically, pre-trained on Kinetics-400~\cite{K400}, our \sexyname brings the gain of 2.3\% on Something-Something V2~\cite{SSV2}, 0.9\% on Kinetics-400, 0.4\% on UCF101~\cite{UCF101}, and 4.7\% on HMDB51~\cite{HMDB51}.

\begin{figure*}[t]
  \centering
   \includegraphics[width=\linewidth]{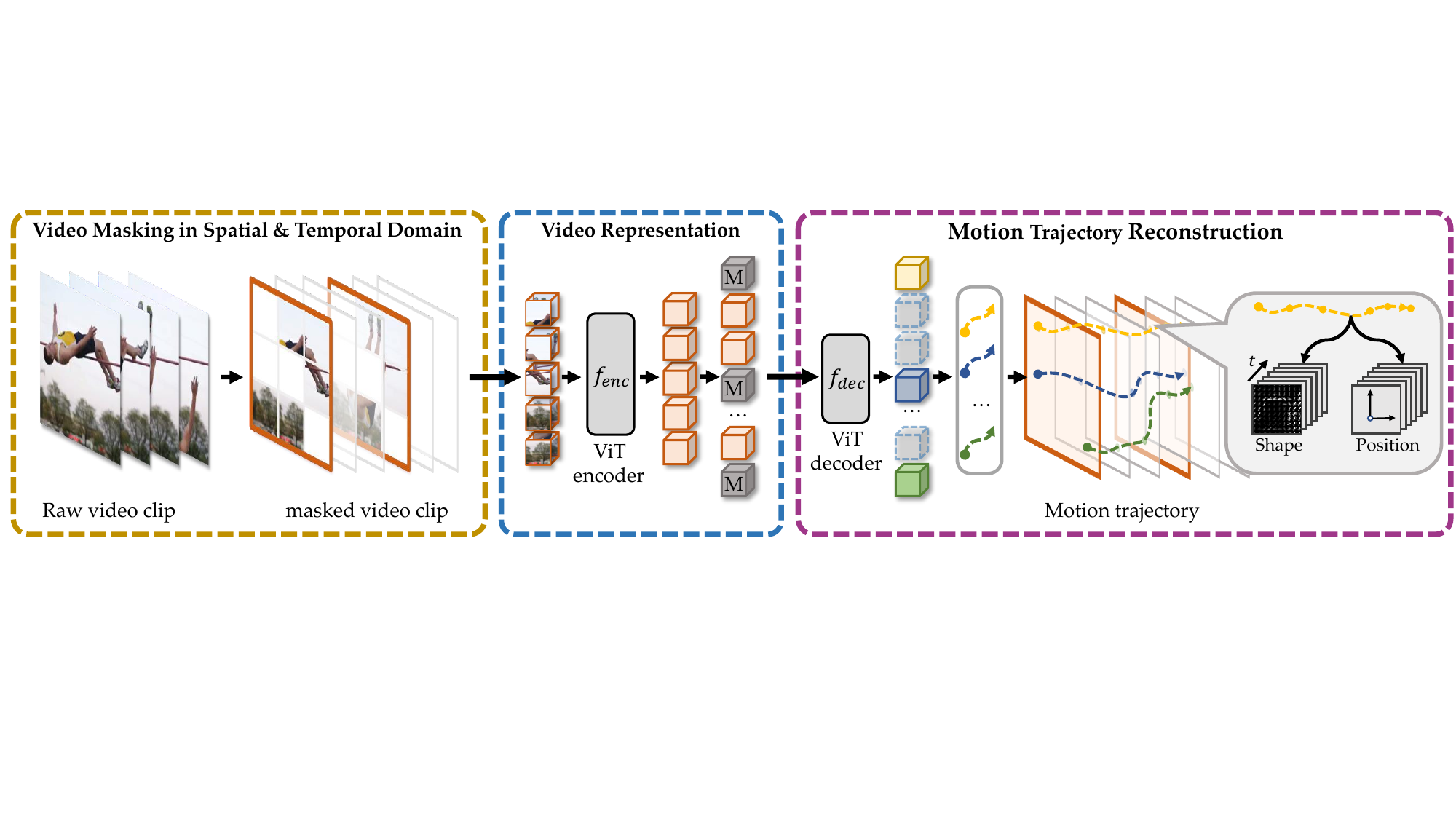}
   \caption{\textbf{Overview of \fullname (\sexyname)}. 
   Given a sparsely sampled video, we first divide it into several patches and randomly mask out some of them. And then, we feed the remaining patches to a ViT encoder to extract video representation. Last, a lightweight ViT decoder is involved to predict the content in the masked region, \ie, a motion trajectory containing position changes and shape changes of moving objects.
   }
   \label{fig:overview}
  \vspace{-4mm}
\end{figure*}

\section{Related Work}
\paragraph{Self-supervised Video Representation Learning.}
Self-supervised video representation learning aims to learn discriminative video features for various downstream tasks in the absence of accurate video labels. To this end, most of the existing methods try to design an advanced pretext task like predicting the temporal order of shuffled video crops~\cite{VCOP}, perceiving the video speediness~\cite{SpeedNet,RSPNet} or solving puzzles~\cite{VCP,3DRoTNet}. In addition, contrastive learning is also widely used in this domain~\cite{CVRL,LSTCL,SVT,BE,ASCNet,CoCLR,liu2021blessings,chen2020look,chen2022weakly}, which constrains the consistency between different augmentation views and brings significant improvement.
In particular, ASCNet and CoCLR ~\cite{ASCNet,CoCLR} focus on mining hard positive samples in different perspectives. 
Optical flow has also been proven to be effective in capturing motion information~\cite{PMAS,ArrowofTime}.
Besides, tracking video objects' movement is also used in self-supervised learning~\cite{Track2015,Track2021,Cycletime,chen2019weakly,chen2021grounding}. Among them, Wang \emph{et al.}~\cite{Cycletime} only utilizes the spatial encoder to extract frame appearance information. CtP framework~\cite{Track2021} and Siamese-triplet network~\cite{Track2015} only require the model to figure out the position and size changes of a specific video patch. Different from these methods, our proposed \sexyname trace the fine-grained movement and shape changes of different parts of objects in the video, hence resulting in a superior video representation. 
Tokmakov \emph{et al.}~\cite{ECCV20WS} utilize Dense Trajectory to provide initial pseudo labels for video clustering. 
However, the model does not predict trajectory motion features explicitly. Instead, we consider long-term and fine-grained trajectory motion features as explicit reconstruction targets. 

\para
\noindent\textbf{Mask Modeling for Vision Transformer.}
Recently, BEiT and MAE~\cite{BEiT,MAE} show two excellent mask image modeling paradigms. BEVT ~\cite{BEVT} and VideoMAE~\cite{VideoMAE} extend these two paradigms to the video domain. 
To learn visual representations, BEVT~\cite{BEVT} predicts the discrete tokens generated by a pre-trained VQ-VAE tokenizer~\cite{VQ-VAE}.
Nevertheless, pre-training such a tokenizer involves an unbearable amount of data and computation. In contrast, VideoMAE~\cite{VideoMAE} pre-train the Vision Transformer by regressing the RGB pixels located in the masked tubes of videos. Due to asymmetric encoder-decoder architecture and extremely high masking ratio, model pre-training is more efficient with VideoMAE. Besides, MaskFeat~\cite{MaskFeat} finds that predicting the Histogram of Gradient (HOG\cite{HOG}) of masked video contents is a strong objective for the mask-and-predict paradigm.
Existing methods only consider static information in each video frame, thus the model can speculate the masked area by watching the visible area in each frame independently and failed to learn \key (cf.  Section~\ref{sec:revisit}). 
Different from the video prediction methods~\cite{STVAE,SSLSTM,Brave,MaskViT} that predict the future frames in pixel or latent space, our \fullname paradigm predicts incoming fine-grained motion in masked video regions, including position changes and shape changes. 

\section{Proposed Method}
We first revisit the current masked video modeling task for video representation learning (cf. Section~\ref{sec:revisit}). 
Then, we introduce our \fullnamelite (\sexyname), where we change the task from recovering appearance to recovering motion trajectory (cf. Section~\ref{sec:mmm}).

\subsection{Rethinking Masked Video Modeling}
\label{sec:revisit}
Given a video clip sampled from a video, self-supervised video representation learning aims to learn a feature encoder $f_{enc}(\cdot)$ that maps the clip to its corresponding feature that best describes the video.
Existing masked video modeling methods~\cite{VideoMAE,MaskFeat} attempt to learn such a feature encoder through a mask-and-predict task. Specifically, the input clip is first divided into multiple non-overlapped 3D patches. Some of these patches are randomly masked and the remaining patches are fed into the feature encoder, followed by a decoder $f_{dec}(\cdot)$ to reconstruct the information in the masked patches. Different works aim to reconstruct different information (\eg, raw pixel in VideoMAE~\cite{VideoMAE} and HOG in MaskFeat~\cite{MaskFeat}). 

However, the existing works share a common characteristic where they all attempt to recover static appearance information of the masked patches. 
Since an image with a high masking ratio (85\% in MAE\cite{MAE}) can be well reconstructed~\cite{MAE,MaskFeat}, we conjecture that the masked appearance information of a video can also be reconstructed frame by frame independently. 
\todo{In this sense, the model may focus more on the contents in the same frame. This may hinder the models from learning important temporal clues, which is critical for video representation. We empirically study this conjecture in the ablation study (cf. Section~\ref{exp:revisit}).}

\subsection{General Scheme of \sexyname}
\label{sec:mmm}
\todo{To better learn temporal clues of a video}, our \sexyname changes the reconstruction content from static appearance information to object motion information, including position and shape change of objects.
As shown in Fig.~\ref{fig:overview}, a video clip is sparsely sampled from a video and is divided into a number of non-overlapped 3D patches of size $t\times h\times w$, corresponding to time, height, and width.
We follow VideoMAE~\cite{VideoMAE} to use the tube masking strategy, where the masking map is the same for all frames, to mask a subset of patches.
For computation efficiency, we follow MAE\cite{MAE} to only feed the unmasked patches (and their positions) to the encoder. The output representation together with learnable $[\mathrm{MASK}]$ tokens are fed to a decoder to reconstruct motion trajectory $\bf z$ in the masked patches. The training loss for MME is 
\begin{equation}
\label{eq:loss}
\mathcal{L}=\sum_{i \in \mathcal{I}} |\bz_i - \hat{\bz}_i|^2,
\end{equation}
where $\hat{\bz}$ is the predicted motion trajectory, and $\mathcal{I}$ is the index set of motion trajectories in all masked patches. 

Motivated by the fact that we humans recognize actions by perceiving position changes and shape changes of moving objects, we leverage these two types of information to represent the motion trajectory.
Through pre-training on MME task, the model is endowed with the ability to explore \key. 
Another important characteristic of the motion trajectory is that it contains fine-grained motion information extracted at raw video rate.
This fine-grained motion information provides the model with a supervision signal to anticipate fine-grained action from sparse video input. 
In the following, we will introduce the proposed motion trajectory in detail.

\begin{table*}[t]
    \centering
    \resizebox{0.9\textwidth}{!}{
    \begin{tabular}[ct]{l c c c c c c c c}
        \toprule
        Method  & Backbone & Pre-train & Label & Epoch & Frames & GFLOPs & Param. & Acc.@1 \\
        \midrule
        TimeSformer~\cite{TimeSformer} & ViT-B & IN-21K & \tiny\Checkmark & 15 & 8 & $196\!\times\!1\!\times\!3$ & 121M & 59.5 \\
        ViViT-L~\cite{ViViT} & ViT-L & IN-21K+K400 & \tiny\Checkmark & 35 & 16 & $3992\times3\times4$ & 311M & 65.4 \\
        Motionformer~\cite{Motionformer} & ViT-B & IN-21K+K400 & \tiny\Checkmark & 35 & 16 & $370\times1\times3$ & 109M & 66.5 \\
        SlowFast~\cite{SlowFast} & ResNet-101 & K400 & \tiny\Checkmark & 196 & 8+32 & $106\times1\times3$ & 53M & 63.1\\
        MViT~\cite{MViT} & MViT-B & K400 & \tiny\Checkmark & 100 & 64 & $455\times1\times3$ & 37M & 67.7 \\
        \midrule
        SVT~\cite{SVT} & ViT-B & IN-21K+K400 & \tiny\XSolidBrush & 20 & 8+64 & $196\!\times\!1\!\times\!3$ & 121M & 59.6 \\
        LSTCL~\cite{LSTCL} & Swin-B & K400 & \tiny\XSolidBrush & 200 & 16 & $360\!\times\!5\!\times\!1$ & 88M & 67.0 \\
        BEVT~\cite{BEVT} & Swin-B & K400+DALL-E & \tiny\XSolidBrush & 150 & 32 & $282\!\times\!1\!\times\!3$ & 88M & 67.1 \\
        VIMPAC~\cite{VIMPAC} & ViT-L & HowTo100M+DALLE & \tiny\XSolidBrush & 100 & 10 & N/A$\times10\times3$ & 307M & 68.1 \\
        VideoMAE~\cite{VideoMAE} & ViT-B & SSv2 & \tiny\XSolidBrush & 400 & 16 & $180\!\times\!2\!\times\!3$ & 87M & 67.6 \\
        VideoMAE~\cite{VideoMAE} & ViT-B & SSv2 & \tiny\XSolidBrush & 800 & 16 & $180\!\times\!2\!\times\!3$ & 87M & 69.3 \\
        VideoMAE~\cite{VideoMAE} & ViT-B & K400 & \tiny\XSolidBrush & 800 & 16 & $180\!\times\!2\!\times\!3$ & 87M & 68.2 \\
        \midrule
        \textbf{\sexyname (Ours)} & ViT-B & SSv2 & \tiny\XSolidBrush & 400 & 16 & $180\!\times\!2\!\times\!3$ & 87M & \textbf{69.2} \\
        \textbf{\sexyname (Ours)} & ViT-B & SSv2 & \tiny\XSolidBrush & 800 & 16 & $180\!\times\!2\!\times\!3$ & 87M & \textbf{70.0} \\
        \textbf{\sexyname (Ours)} & ViT-B & K400 & \tiny\XSolidBrush & 400 & 16 & $180\!\times\!2\!\times\!3$ & 87M & \textbf{69.3} \\
        \textbf{\sexyname (Ours)} & ViT-B & K400 & \tiny\XSolidBrush & 800 & 16 & $180\!\times\!2\!\times\!3$ & 87M & \textbf{70.5} \\
        \bottomrule
    \end{tabular}}
    \caption{\small\textbf{Comparison with state-of-the-arts on Something-Something V2.} Our \sexyname outperforms the previous best method. Remarkably, \sexyname show better transferability ability (transfer from K400 to SSv2) compared with VideoMAE.}
    \label{tab:sota-ssv2}
\end{table*} 

\begin{table*}[t]
    \centering
    \resizebox{0.9\textwidth}{!}{
    \begin{tabular}[ct]{l c c c c c c c c c}
        \toprule
        Method  & Backbone & Pre-train & Label & Epoch & Frames & GFLOPs & Param. & Acc.@1 \\
        \midrule
        TimeSformer~\cite{TimeSformer} & ViT-B & IN-21K & \tiny\Checkmark & 15 & 8 & $196\!\times\!1\!\times\!3$ & 121M & 78.3 \\
        Motionformer~\cite{Motionformer} & ViT-B & IN-21K & \tiny\Checkmark & 35 & 16 & $370\!\times\!10\!\times\!3$ & 121M & 79.7 \\
        Motionformer~\cite{Motionformer} & ViT-HR & IN-21K & \tiny\Checkmark & 35 & 16 & $959\!\times\!10\!\times\!3$ & 382M & 81.1 \\
        \midrule
        VIMPAC~\cite{VIMPAC} & ViT-L & HowTo100M+DALLE & \tiny\XSolidBrush & 100 & 10 & N/A$\times10\times3$ & 307M & 77.4 \\
        SVT~\cite{SVT} & ViT-B & IN-21K+K400 & \tiny\XSolidBrush & 20 & 8+64 & $196\!\times\!1\!\times\!3$ & 121M & 78.1 \\
        BEVT~\cite{BEVT} & Swin-B & K400+DALL-E & \tiny\XSolidBrush & 150 & 32 & $282\!\times\!1\!\times\!3$ & 88M & 76.2 \\
        BEVT~\cite{BEVT} & Swin-B & \scriptsize IN21K+K400+DALL-E & \tiny\XSolidBrush & 150 & 32 & $282\!\times\!1\!\times\!3$ & 88M & 80.6 \\
        LSTCL~\cite{LSTCL} & Swin-B & K400 & \tiny\XSolidBrush & 200 & 16 & $360\!\times\!5\!\times\!1$ & 88M & 81.5 \\
        OmniMAE~\cite{OmniMAE} & ViT-B & IN1K+K400 & \tiny\XSolidBrush & 1600 & 32 & $180\!\times\!5\!\times\!3$ & 87M & 80.6 \\
        VideoMAE~\cite{VideoMAE} & ViT-B & K400 & \tiny\XSolidBrush & 1600 & 16 & $180\!\times\!5\!\times\!3$ & 87M & 80.9 \\
        ST-MAE~\cite{MAE-v} & ViT-B & K400 & \tiny\XSolidBrush & 1600 & 16 & $180\!\times\!7\!\times\!3$ & 87M & 81.3 \\
        \midrule
        \textbf{\sexyname (Ours)} & ViT-B & K400 & \tiny\XSolidBrush & 1600 & 16 & $180\!\times\!7\!\times\!3$ & 87M & \textbf{81.8} \\
        
        \bottomrule
    \end{tabular}}
    \caption{\small\textbf{Comparison with state-of-the-arts on Kinetics-400.} Our \sexyname outperforms the previous best method.}
    \vspace{-2mm}
    \label{tab:sota-k400}
\end{table*} 

\subsection{Motion Trajectory for MME}
\label{sec:motion-trajectory}
The motion of moving objects can be represented in various ways such as optical flow~\cite{Farneback}, histograms of optical flow (HOF)~\cite{HOF}, and motion boundary histograms (MBH)~\cite{HOF}. However, these descriptors can only represent short-term motion between two adjacent frames. 
We hope our motion trajectory represents long-term motion, which is critical for video representation. 
To this end, inspired by DT~\cite{dt}, we first track the moving object in the following $L$ frames to cover a longer range of motion, resulting in a trajectory $\bT$, \ie, 
\begin{equation}
\label{eq:trajectory}
    \bT = (\bp_t, \bp_{t+1}, \cdots, \bp_{t+L}),
\end{equation}
where $\bp_t = (x_t, y_t)$ represents a point located at $(x_t, y_t)$ of frame $t$, and $(\cdot, \cdot)$ indicates the concatenation operation.
Along this trajectory, we fetch the position features $\bz^p$ and shape features $\bz^s$ of this object to compose a motion trajectory $\bz$, \ie, 
\begin{equation}
\bz = (\bz^p, \bz^s).
\label{eq:motion-trajectory}
\end{equation}
The position features are represented by the position transition relative to the last time step, while the shape features are the HOG descriptors of the tracked object in different time steps.

\para
\noindent\textbf{Tracking objects using spatially and temporally dense trajectories.}
Some previous works~\cite{ObjTraj,chencomphy,ding2021dynamic,chen2021grounding} try to use one trajectory to represent the motion of an individual object. In contrast, DT~\cite{dt} points out that tracking spatially dense feature points sampled on a grid space performs better since it ensures better coverage of different objects in a video. Following DT~\cite{dt}, we use spatially dense grid points as the initial position of each trajectory. 
Specifically, we uniformly sample $K$ points in a masked patch of size $t\times h\times w$, where each point indicates a part of an object. For each point, we track it through temporally dense $L$ frames according to the dense optical flow, resulting in $K$ trajectories. 
In this way, the model is able to capture spatially and temporally dense motion information of objects through the mask-and-predict task.

As a comparison, the reconstruction contents in existing works~\cite{VideoMAE,MaskFeat} often extract temporally sparse videos sampled with a large stride $s>1$.
The model takes as input a sparse video and predicts these sparse contents for learning video representation. Different from these works, our model also takes as input sparse video but we push the model to interpolate motion trajectory containing fine-grained motion information. 
This simple trajectory interpolation task does not increase the computational cost of the video encoder but helps the model learn more fine-grained action information even given sparse video as input. 
More details about dense flow calculating and trajectory tracking can be found in Appendix.

\para
\noindent\textbf{Representing position features.}
Given a trajectory $\bT$ consisting of the tracked object position at each frame, we are more interested in the related movement of objects instead of their absolute location. 
Consequently, we represent the position features with related movement between two adjacent points $\Delta \bp_t = \bp_{t+1} - \bp_t$, \ie, 
\begin{equation}
    \bz^p = (\Delta \bp_t,...,\Delta \bp_{t+L-1}),
\end{equation}
where $\bz^p$ is a $L \times 2$ dimensional feature. 
As each patch contains $K$ position features, we concatenate and normalize them as position features part of the motion trajectory.

\para
\noindent\textbf{Representing shape features.}
Besides embedding the movement, the model also needs to be aware of the shape changes of objects to recognize actions. Inspired by HOG~\cite{HOG}, we use histograms of oriented gradients (HOG) with 9 bins to describe the shape of objects. 

Compared with existing works~\cite{MaskFeat,VideoMAE} that reconstruct HOG in every single frame, we are more interested in the dynamic shape changes of an object, which can better represent action in a video. To this end, we follow DT~\cite{dt} to calculate trajectory-aligned HOG, consisting of HOG features around all tracked points in a trajectory, \ie,
\begin{equation}
    \bz^s = ({\rm HOG} (\bp_t),...,{\rm HOG} (\bp_{t+L-1})),
\end{equation}
where ${\rm HOG}(\cdot)$ is the HOG descriptor and $\bz^s$ is a $L \times 9$ dimensional feature. 
Also, as one patch contains $K$ trajectories, we concatenate $K$ trajectory-aligned HOG features and normalize them to the standard normal distribution as the shape features part of the motion trajectory.

\section{Experiments}
\label{exps}


\para
\noindent\textbf{Implementation details.}
We conduct experiments on Kinetics-400 (K400), Something-Something V2 (SSV2), UCF101 and HMDB51 datasets.
Unless otherwise stated, we follow previous trails~\cite{VideoMAE} and feed the model a 224$\times$224 16-frame clip with strides 4 and 3 for K400 and SSV2 respectively. We use vanilla ViT-Base~\cite{vit} as the backbone in all experiments. As for the motion trajectory, we set the trajectory length to $L=6$ and the number of trajectories  in each patch to $K=4$. Our models are trained on 16 NVIDIA GeForce RTX 3090 GPUs. 
Note that for all the ablation studies, we use a smaller pre-training dataset and reduce the input patches of fine-tuning to speed up the experiments, details can be found in Appendix.


\subsection{Comparisons with State-of-the-arts}

\paragraph{Fine-tuning.} 
On the temporal-heavy dataset SSv2, we conduct experiments using two settings, namely in-domain pre-training (\ie, pre-train and fine-tune on the same dataset) and cross-domain pre-training (\ie, pre-train and fine-tune on different datasets). In Tab.~\ref{tab:sota-ssv2}, we have two main observations. 1) Our \sexyname significantly performs better than previous state-of-the-arts VideoMAE~\cite{VideoMAE} on both in-domain and cross-domain settings. Specifically, under cross-domain settings (transferring from K400 to SSv2) with 800 epochs, our \sexyname outperforms it by 2.3\%, increasing the accuracy from 68.2\% to 70.5\%. 2) Our \sexyname performs better under cross-domain setting compared with in-domain setting, which is contrary to the existing observation on VideoMAE~\cite{VideoMAE} and ST-MAE~\cite{MAE-v}. We attribute it to our \fullnamelite pretext task that helps models focus more on learning temporal clues instead of learning domain-specific appearance information (\eg, color, lighting). The learned temporal clues are more universal across different domains for video representation. This also shows the potential for taking advantage of larger datasets for pre-training, which we leave for future work.

In Tab.~\ref{tab:sota-k400}, our \sexyname also achieves state-of-the-art performance on another large-scale dataset K400 with a gain of 0.5\%. In Tab.~\ref{tab:sota-ucf}, we also evaluate the transferability of \sexyname by transferring a K400 pre-trained model to two small-scale datasets, namely UCF101 and HMDB51. Our \sexyname achieves a large gain of 4.7\% on HMDB51 and 0.4\% on UCF101 compared with VideoMAE. This further demonstrates the strong ability of \sexyname to learn from the large-scale dataset and transfer it to a small-scale one.

\begin{table}[t]
    \centering
    \small
    \resizebox{0.48\textwidth}{!}{
    \begin{tabular}{ccccc}
        \toprule
        Method    & Backbone & Pre-train & UCF101 & HMDB51 \\
        \midrule
        XDC~\cite{XDC}        & R(2+1)D  & IG65M     & 94.2   & 67.1   \\
        GDT~\cite{GDT}        & R(2+1)D  & IG65M     & 95.2   & 72.8   \\
        CVRL~\cite{CVRL}       & R50      & K400      & 92.9   & 67.9   \\
        CORP$_f$~\cite{CORPf}   & R50      & K400      & 93.5   & 68.0   \\
        $\rho$BYOL~\cite{RhoBYOL} & R50      & K400      & 94.2   & 72.1   \\
        VideoMAE~\cite{VideoMAE}   & ViT-B    & K400      & 96.1   & 73.3   \\
        \midrule
        \textbf{\sexyname (Ours)} & ViT-B & K400 & \textbf{96.5} & \textbf{78.0} \\
        \bottomrule
    \end{tabular}
    }
    \caption{\small\textbf{Comparison with state-of-the-arts on UCF101 and HMDB51}.}
    \label{tab:sota-ucf}
    \vspace{-3mm}
\end{table}

\para
\noindent\textbf{Linear probing.} 
\label{exp:linear}
We further evaluate our \sexyname under liner probing setting on the SSv2 dataset as this dataset focuses more on the action. 
We follow SVT~\cite{SVT} to fix the transformer backbone and train a linear layer for 20 epochs. In Tab.~\ref{tab:linear-ssv2}, our method significantly outperform previous contrastive-based method~\cite{SVT} by 10.9\% and masked video modeling methods~\cite{VideoMAE} by 4.9\%. These results indicate that the representation learned from our \sexyname task contains more motion clues for action recognition.


\subsection{Ablation Study on Reconstruction Content}
In this section, we investigate the effect of different reconstruction contents, including appearance, short-term motion, and the proposed long-term fine-grained motion.

\subsubsection{Evaluating Appearance Reconstruction}
\label{exp:revisit}
\begin{figure}[t]
    \centering
    \includegraphics[width=0.47\textwidth]{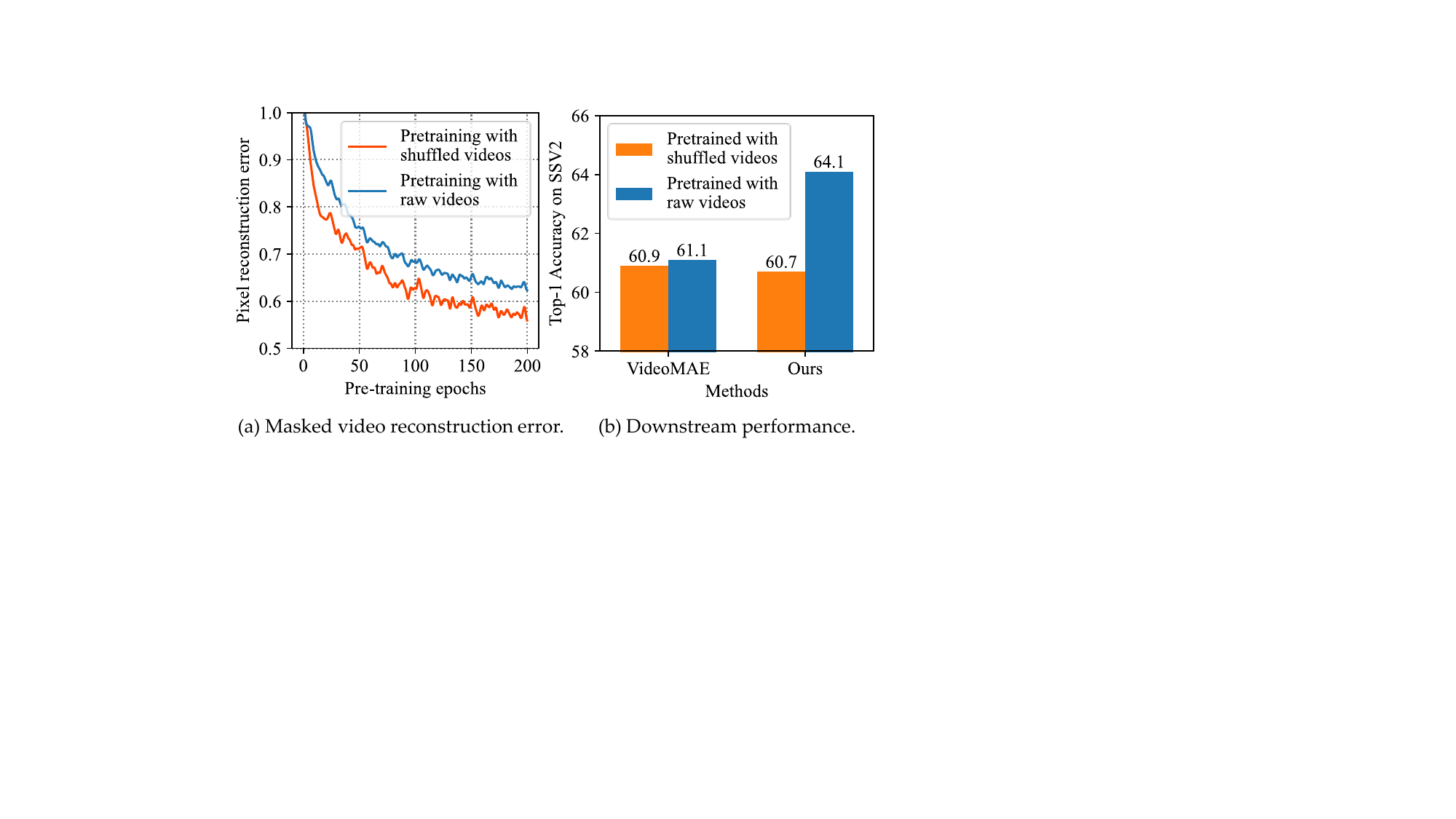}
    \caption{We perform mask-and-predict pre-training using temporal shuffled videos and raw videos and then transfer them to the downstream task. (a) Masked appearance (\ie pixels) is well reconstructed without temporal information. (b) Our \sexyname outperforms VideoMAE when both provided temporal information.}
    \label{fig:toy-exp}
    \vspace{-2mm}
\end{figure}

\mk{In section~\ref{sec:revisit}, we conjecture that appearance reconstruction is finished in each video frame independently. To verify this conjecture, we corrupt the temporal information of a video by randomly shuffling its frames.
\cph{Since the video frames are randomly permuted, it is hard for a model to leverage information from other frames to reconstruct the current frame through their temporal correlation.} We mask 90\% regions of these videos following VideoMAE~\cite{VideoMAE}. In Fig.~\ref{fig:toy-exp} (a), we find that the reconstruction error (\ie, L2 error between predicted and ground-truth pixel) convergences to a low value. We also conduct the same experiment using raw video without shuffling. \cph{The model undergoes a similar convergence process.} This demonstrates that the masked pixels are well reconstructed without temporal information.}


To further evaluate whether the model pre-trained on the appearance reconstruction task (\ie, VideoMAE) can well capture \key, we transfer two VideoMAE models (pre-trained on shuffled videos and raw videos, respectively) to the downstream action recognition task. In Fig.~\ref{fig:toy-exp} (b), these two models perform competitively, indicating that removing the temporal information from the pre-training videos barely affects the learned video representation. We speculate this is because the VideoMAE model pays little attention to temporal information when performing the appearance reconstruction pre-training task. As a comparison, our \sexyname paradigm performs significantly better than VideoMAE when the temporal information is provided for pre-training (64.1\% \vs 60.9\%). Once the temporal information is removed, \sexyname performs similarly with VideoMAE (60.7\% \vs 60.9). This demonstrates our \sexyname paradigm takes great advantage of temporal information in the pre-training phase.

\begin{table}[t]
    \centering
    \small
    \begin{tabular}{lcccc}
    \toprule
         Method & Backbone & Pre-train  & Acc.@1 \\
         \midrule
         TimeSformer~\cite{TimeSformer} & ViT-B & IN-21K & 14.0 \\
         SVT~\cite{SVT} & ViT-B & IN-21K+K400 & 18.3 \\
         VideoMAE~\cite{VideoMAE} & ViT-B & SSv2 & 24.3 \\
         \midrule
         \textbf{\sexyname (Ours)} & ViT-B & SSv2 & \textbf{29.2} \\
         \bottomrule
    \end{tabular}
    \caption{\textbf{Linear Probing on Something-Something V2}.}
    \label{tab:linear-ssv2}
    \vspace{-2mm}
\end{table}

\subsubsection{Evaluating Motion Reconstruction}

In the following ablations, we separately verify the effectiveness of two proposals of our methods:

\begin{table}[t]
    \centering
    \small
    \begin{tabular}{l|cc}
         \toprule
         Reconstruction content & Acc.@1 & Acc.@5 \\
         \midrule
         Pixel & 61.1 & 86.6 \\
         HOG & 61.1 & 86.9 \\
         \midrule
         HOG + Flow & 60.9 & 86.4  \\
         HOG + HOF & 61.3 & 86.6 \\
         HOG + MBH & 61.4 & 86.8 \\
         \midrule
         HOG + Traj. w/o shape & 63.5 & 88.2 \\ 
         \cellcolor{mygray}Traj. (ours) & \cellcolor{mygray}\textbf{64.1} & \cellcolor{mygray}\textbf{88.4} \\
         \bottomrule
    \end{tabular}
    \caption{\textbf{Comparison of different Reconstruction content.} Our motion trajectory outperforms all the other contents.}
    \label{tab:motion-target}
    \vspace{-2mm}
\end{table} 

\para
\noindent\textbf{Effectiveness of motion trajectory.}
We compare our motion trajectory with two appearance reconstruction baselines and three short-term motion reconstruction baselines.

\begin{itemize}
\setlength{\itemsep}{2pt}
\item \textbf{Appearance reconstruction baselines}: 1) \textbf{HOG}: predicting HOG features~\cite{HOG} in the middle frame of each 3D patch~\cite{MaskFeat}. 2) \textbf{Pixel}: predicting all the pixels of each 3D patch~\cite{MAE,VideoMAE,MAE-v}.

\item \textbf{Short-term motion reconstruction baselines}: 
1)  \textbf{Flow}: predicting the dense optical flow map of the masked patches. 
2) \textbf{HOF}: predicting the histogram of optical flow orientation. 
3) \textbf{MBH}: predicting the motion boundary histogram, which is the second-order local statistic of optical flow. 
\end{itemize}


\label{exp:feature-combination}

In Tab.~\ref{tab:motion-target}, two types of appearance reconstruction contents (\ie, Pixel and HOG) perform similarly. Based on the HOG baseline, predicting additional short-term motion contents brings slight improvement (at least by 0.2\%) except for directly predicting optical flow (decreases by 0.2\%). 
We suspect that predicting the dense flow map for each pixel is too hard for a model and thus harms the performance. 
Instead, the histogram version of flow (\ie, HOF and MBH) help to reduce the prediction ambiguity~\cite{MaskFeat}, resulting in better performance.
Furthermore, after adding our slight variant of motion trajectory (\ie, with position features but without shape features shown in Eq.~(\ref{eq:motion-trajectory})), the performance improves significantly compared to both appearance reconstruction baselines (by 2.4\%) and short-term motion reconstruction baselines (by 2.1\%). On top of this variant, adding trajectory-aligned shape features further brings additional gain by 0.6\%. \rebb{Because tracking the trajectory of objects in multiple frames provides long-term motion compared to short-term motion reconstruction, we believe the learning of \textbf{long-term motion} clues is one of the keys to enhancing video representation.}

\begin{table}[t]
    \centering
    \small
    \begin{tabular}{lc|cc}
         \toprule
         Reconstruction content & Interpolation & Acc.@1 & Acc.@5 \\
         \midrule
         HOG & \tiny\XSolidBrush & 61.1 & 86.9 \\
         HOG & \tiny\Checkmark & 60.9 & 86.8 \\
         \midrule
         HOG + MBH & \tiny\XSolidBrush & 61.3 & 86.8 \\
         HOG + MBH & \tiny\Checkmark & 61.4 & 86.8 \\
         \midrule
         Traj. (ours) & \tiny\XSolidBrush & 62.2 & 87.5 \\
         \cellcolor{mygray}Traj. (ours) & \cellcolor{mygray}\tiny\Checkmark & \cellcolor{mygray}\textbf{64.1} & \cellcolor{mygray}\textbf{88.4} \\
         \bottomrule
    \end{tabular}
    \caption{\textbf{Effectiveness of trajectory interpolation.} Interpolating the motion trajectory brings significant improvement compared to the baselines.}
    \vspace{-2mm}
    \label{tab:action-super}
\end{table}

\begin{table*}[t]
    \small
    \begin{subtable}[t]{0.27\textwidth}
    \vspace{0pt}
    \centering
    \makeatletter\def\@captype{table}
       \begin{tabular}{ccc}
           \toprule
           Length & Acc.@1 & Acc.@5\\
           \midrule
           $L=4$ & 63.2 & 88.0 \\
           \cellcolor{mygray}$L=6$ & \cellcolor{mygray}\textbf{64.1} & \cellcolor{mygray}\textbf{88.4} \\
           $L=8$ & 63.3 & 88.0 \\
           \bottomrule
        \end{tabular}
        \vspace{2pt}
        \caption{\small\textbf{Effect of trajectory length}.}
        \label{tab:traj-len}
    \end{subtable}
    \hfill
    \begin{subtable}[t]{0.3\textwidth}
    \vspace{0pt}
    \centering
    \makeatletter\def\@captype{table}
       \begin{tabular}{ccc}
           \toprule
           Setting & Acc.@1 & Acc.@5 \\
           \midrule
           \specialrule{0em}{2.2pt}{2.2pt}
           w/o patch norm. & 62.1 & 87.44 \\
           \specialrule{0em}{2.2pt}{2.2pt}
           \cellcolor{mygray}w/ patch norm. & \cellcolor{mygray}\textbf{64.1} & \cellcolor{mygray}\textbf{88.4}\\
           \specialrule{0em}{2.2pt}{2.2pt}
           \bottomrule
        \end{tabular}
        \vspace{2pt}
        \caption{\small\textbf{Effect of trajectory normalization}.
        }
        \label{tab:traj-norm}
    \end{subtable}
    \hfill
    \begin{subtable}[t]{0.38\textwidth}
    \vspace{0pt}
    \centering
    \makeatletter\def\@captype{table}
       \begin{tabular}{ccc}
           \toprule
           Trajectory density  & Acc.@1 & Acc.@5 \\
           \midrule
           \cellcolor{mygray}Dense & \cellcolor{mygray}\textbf{64.1} & \cellcolor{mygray}\textbf{88.4}\\
           Sparse (Max) & 60.2 & 86.3 \\
           Sparse (Mean) & 61.8 & 87.1 \\
           \bottomrule
        \end{tabular}
        \vspace{2pt}
        \caption{\small\textbf{Effect of trajectories density}. 
        }
        \label{tab:traj-density}
    \end{subtable}
    \hfill
    
    \begin{subtable}[t]{0.27\textwidth}
    \vspace{12pt}
    \centering
    \makeatletter\def\@captype{table}
        \begin{tabular}{ccc}
           \toprule
           Stride & Acc.@1 & Acc.@5 \\
           \midrule
           $s=4$ & 62.9 & 87.7 \\
           \cellcolor{mygray}$s=3$ & \cellcolor{mygray}\textbf{64.1} & \cellcolor{mygray}\textbf{88.4} \\
           $s=2$ & 63.5 & 88.3 \\
           $s=1$ & 62.7 & 87.7\\
           \bottomrule
        \end{tabular}
        \vspace{4pt}
        \caption{\small\textbf{Effect of sampling stride}.
        }
       \label{tab:temporal-masking}
    \end{subtable}
    \hfill
    \begin{subtable}[t]{0.3\textwidth}
    \vspace{12pt}
    \centering
    \makeatletter\def\@captype{table}
        \begin{tabular}{cccc}
           \toprule
           Ratio & Type & Acc.@1 & Acc.@5 \\
           \midrule
           90\% & Tube & 62.7 & 87.8 \\
           \cellcolor{mygray}70\% & \cellcolor{mygray}Tube & \cellcolor{mygray}\textbf{64.1} & \cellcolor{mygray}\textbf{88.4} \\
           40\% & Tube & 61.3 & 86.9\\
           40\% & Cube & 61.2 & 86.9 \\
           \bottomrule
        \end{tabular}
        \vspace{4pt}
        \caption{\small\textbf{Effect of spatial masking}. 
        }
        \label{tab:spatial-masking}
    \end{subtable}
    \hfill
    \begin{subtable}[t]{0.38\textwidth}
    \vspace{9pt}
    \centering
    \makeatletter\def\@captype{table}
        \begin{tabular}{ccccc}
           \toprule
           Type & Depth & FLOPs & Acc.@1 & Acc.@5 \\
           \midrule
           \multirow{2}*{Joint} & \cellcolor{mygray}1 & \cellcolor{mygray}51G & \cellcolor{mygray}\textbf{64.1} & \cellcolor{mygray}\textbf{88.4}\\
           ~ & 2 & 55G & 63.4 & 88.3 \\
           \midrule
           \multirow{2}*{Sep.} & 1 & 56G & 63.4 & 88.4 \\
           ~ & 2 & 65G & 63.3 & 88.0 \\
           \bottomrule
        \end{tabular}
        \vspace{2pt}
        \caption{\small\textbf{Effect of decoder}. Sep. means two decoders. 
        } 
        \label{tab:decoder-setting}
    \end{subtable}
    \caption{\textbf{Pre-training details.} We explore the detailed setting of our \sexyname pre-training. The default entry is marked in \colorbox{mygray}{\color{black}gray}.}
    \label{tab:traj-implement}
    \vspace{-2mm}
\end{table*}

\para
\noindent\textbf{Effectiveness of trajectory interpolation.}
In \sexyname, we feed temporally sparse video to the model and push it to interpolate temporally dense trajectories. We are interested in whether interpolating appearance (\ie, HOG) or short-term motion contents (\ie, MBH) benefit learning fine-grained motion details. To this end, we conduct experiments to predict dense HOG and MBH, \ie, predicting these features of all frames within a masked patch even though these frames are not sampled as input. In Tab.~\ref{tab:action-super}, interpolating HOG or MBH brings little improvement. We suspect this is because these dense features represent the continuous movement of objects poorly, as they have not explicitly tracked the moving objects. In contrast, our temporally dense motion trajectory tracks the object densely, providing temporal fine-grained information for learning. The trajectory interpolation improves our \sexyname by 1.9\%. \rebb{This result proves that the learning of \textbf{fine-grained motion} details is another key to enhancing video representation.}

\subsection{Ablation Study on \texorpdfstring{\sexyname}{Lg} Design}
In this section, we conduct ablation studies to investigate the effect of different parameters for MME pre-training.

\paragraph{Trajectory length.}
\label{exp:traj-len}
Trajectory length $L$ indicates how many frames we track an object for extracting the motion trajectory. If $L=0$, our method degrades to MaskFeat~\cite{MaskFeat}, which only predicts HOG features in the masked patch. In Tab.~\ref{tab:traj-len}, \sexyname achieves the best performance when $L=6$. 
This is because a longer trajectory provides more motion information for learning. But if the trajectory is too long, the drift of flow introduces accumulated noise when tracking the objects, which harms the performance. 


\para
\noindent\textbf{Trajectory density.}
\label{exp:traj-density} 
In our experiments, we spatially dense track $K=4$ motion trajectories. We conduct experiments to evaluate whether it is necessary to densely track multiple trajectories in a patch. Specifically, we assume only one object exists in a patch, and thus the 4 trajectories are similar. We merge these 4 trajectories as one by averaging or selecting the most salient one. We then consider the merged trajectories as the trajectory motion. In Tab.~\ref{tab:traj-density}, using dense trajectories significantly outperform the merged baselines. We suspect the spatial dense trajectories provide more motion information for learning.

\para
\noindent\textbf{Trajectory normalization.}
Due to the irregular distribution of motion and shape information in the video, the trajectory values in different patches vary a lot. To help the model better model the trajectory, we adopt the patch norm method to normalize the trajectory values in each patch into a standard normal distribution, as has also been done in MAE~\cite{MAE}. The mean and variance are calculated by all $K=4$ trajectories in a patch. Tab.~\ref{tab:traj-norm} shows the effectiveness of adopting the patch norm.

\para
\noindent\textbf{Sampling stride.}
In order to control the difficulty of interpolating the motion trajectory, we ablate different temporal sampling strides leading to different levels of the sparse input video. With a medium sampling stride ($s=2$), our \sexyname achieves the best performance. We also found that the performance decreases when the input video shares the same stride as the dense motion trajectory ($s=1$). One possible reason is that it is difficult for the model to anticipate fine-grained motion details in this case.

\para
\noindent\textbf{Spatial masking strategy.}
\label{exp:mask}
We ablated study different spatial masking strategies and ratios in Tab.~\ref{tab:spatial-masking}. We start from the default setting of MaskFeat~\cite{MaskFeat} that uses a cube masking strategy with a 40\% masking ratio, then we increase the masking ratio and adopt the tube masking strategy proposed by VideoMAE~\cite{VideoMAE}. Using tube masking with a medium masking ratio (70\%) performs the best.

\para
\noindent\textbf{Decoder setting.}
We use several transformer blocks as decoder to reconstruct motion trajectory. 
We ablate different architectures of the decoder in Tab.~\ref{tab:decoder-setting}. Using a slight decoder with 1 transformer block performs the best, with the lowest computational cost. Increasing the depth of the decoder or using two parallel decoders to separately predict the position and shape features introduces more computational cost but brings little improvement.

\begin{figure}
    \centering
    \includegraphics[width=0.44\textwidth]{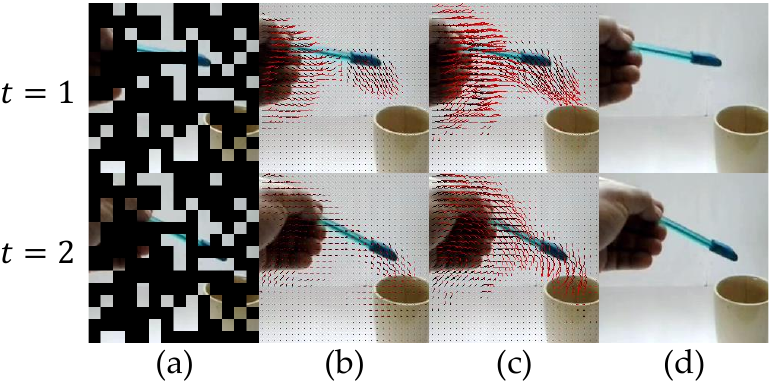}
    \caption{\textbf{Visualization of predicted motion trajectory}. Each column represents (from left to right): (a) Masked frames; (b) ground truth; (c) prediction; (d) original frame. 
    }
    \label{fig:visual}
    \vspace{-4mm}
\end{figure}

\subsection{Visualization Results}
We visualize the predicted motion trajectory in Fig.~\ref{fig:visual}. The model is able to reconstruct the spatial and temporally dense trajectories of different object parts from a temporally sparse input video. Even without seeing the object in the current frame (the hand being masked out), the model is able to locate it through contextual patches and accurately estimate its trajectory. 
We only visualize the position changes since our prediction of HOG features has similar quality to the previous work~\cite{MaskFeat}. 

\section{Conclusion}
We present \fullname, a simple mask-and-predict paradigm that learns video representation by reconstructing the motion contents of masked regions.
In particular, we reconstruct motion trajectory, which tracks moving objects densely to record their position and shape changes.
By reconstructing this motion trajectory, the model learns long-term and fine-grained motion clues from sparse input videos. Empirical results show that our method outperforms the state-of-the-art mask-and-predict methods on Something-Something V2, Kinetics-400, UCF-101, and HMDB-51 datasets for action recognition. 

\section*{Acknowledgement}
This work was partially supported by the National Natural Science Foundation of China (NSFC) (No. 62072190, 62072186), 
Ministry of Science and Technology Foundation Project (No. 2020AAA0106900), 
Program for Guangdong Introducing Innovative and Enterpreneurial Teams (No. 2017ZT07X183), 
Guangdong Basic and Applied Basic Research Foundation (No. 2019B1515130001).

{\small
\bibliographystyle{ieee_fullname}
\bibliography{egbib}
}

\clearpage
\appendix
\begin{leftline}
	{
		\LARGE{\textsc{Appendix}}
	}
\end{leftline}

\section{More Details on Optical Flow Extraction}
\label{app:flow-extraction}
Before generating the trajectory, we need to calculate dense optical flows frame-by-frame to provide the fine-grained motion of each pixel. We follow the previous work to remove camera motion and pre-extract the optical flows of each video in the dataset offline. Details are as below. 

\para
\noindent\textbf{Removing camera motion.} To alleviate the influence of camera motion, we follow previous work~\cite{iDT} to warp the optical flow, which is also be applied in two stream video action recognition method~\cite{TSN}. Specifically, by matching SURF~\cite{SURF} interest point in two frames to estimate the camera motion using RANSC~\cite{RANSAC} algorithm. Then the camera motion is removed by rectifying the frame, and the optical flow is re-calculated using the warped frame. This also relieves the model from being disturbed by the background and makes it pay more attention to the moving objects in the foreground.

\para
\noindent\textbf{Pre-extracting optical flow.} We use the \emph{denseflow}\footnote{\url{https://github.com/yjxiong/dense_flow}} tool box to pre-extract warp optical flow before training following~\cite{iDT,TSN}. We set the upper bound of flow value to 20. The stride of flow is set to $s_{flow}=1$ if perform motion target interpolation, otherwise is equal to the sampling stride of the input rgb frames $s_{flow}=s_{rgb}$. The whole video dataset is split into chunks and processed on 4 nodes, each node is equipped with 2 Intel Xeon CPUs (32 cores per CPU) for video codec and 8 TITAN X GPUs for optical flow calculation speed-up. The extraction process spends about 1 day on the Kinetics400 dataset.

\section{More Details on Motion Trajectory Generation}
\begin{figure}[h]
    \centering
    \makeatletter\def\@captype{figure}
    \includegraphics[width=1\linewidth]{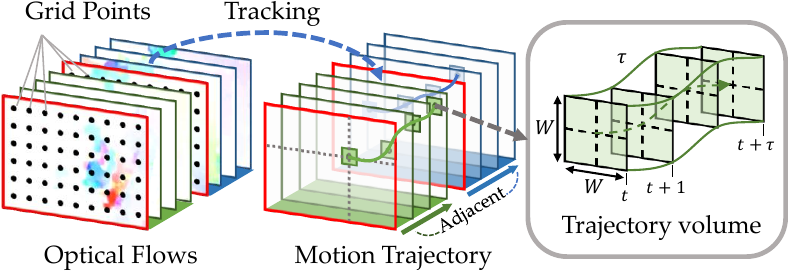}
    \caption{\textbf{The detailed generation process of motion trajectory.} }
    \label{fig:details}
    \vspace{-4mm}
\end{figure}

As mentioned in Section 3, we use a motion trajectory to represent long-term and fine-grained motion, which consists of position features $\bz^p$ and shape features $\bz^s$. To extract these two features, we first need to calculate a continuous trajectory that tracks the position transition of a specific grid point. We then consider the surrounding area of this trajectory as a $\tau\times W\times W$ volume, as shown in Fig.~\ref{fig:details}. The generation process of this trajectory is described in detailed as following.

\label{app:traj-generation}
\para
\noindent\textbf{Tracking the trajectory.}
With the pre-extracted warp optical flow $\omega$, we are able to track the position transition of each densely sampled grid point to produce a continuous trajectory, \ie,
\begin{equation}
\label{eq:tracking}
\bp_{t+1}=(x_{t+1},y_{t+1})=(x_t,y_t)+(M\ast\omega_t)_{(x_t,y_t)},
\end{equation}
where $\omega_t$ is the optical flow at step $t$. $M$ is the median filter kernel shapes $3\times3$ and $\ast$ denotes the convolution operator.

\para
\noindent\textbf{Pre-extracting the trajectory.}
To reduce the io overhead of loading and decoding flow images during pre-training, we pre-process the dataset to generate dense trajectories of each video and pack them in the compressed pickle files. In pre-training, we use a special dataloader to sample $K$ trajectories with length $L$ starting from the first frame of each input 3D patch. During pre-training, we then crop and resize the trajectories in the spatial dimension to maintain the corresponding area with the input RGB frames.

\begin{figure}[ht]
    \centering
    \includegraphics[width=1\linewidth]{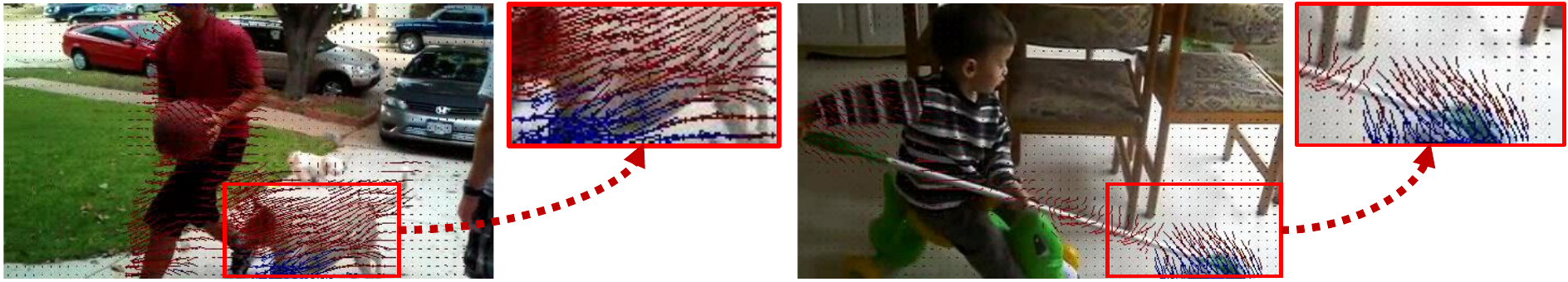}
    \caption{\textbf{Illustration of the inaccurate trajectories} (marked as blue), which are caused by objects moving out of the camera FOV.}
    \label{fig:app-outlier}
    \vspace{-4mm}
\end{figure}

\para
\noindent\textbf{Masking inaccurate trajectory.} We notice that in most of the video samples, the objects move out of the range of the camera FOV. This circumstance often occurs on the objects that are at the edge of the video. The tracked trajectories, in this case, could be inaccurate, as visualized in Fig.~\ref{fig:app-outlier}. We use a loss mask to remove these trajectories from the loss calculation:

\begin{equation}
    \mathcal{L}^M=\frac{1}{N_p}\sum_{i=1}^{N_p}{m_i\cdot\mathcal{L}_i},\ \ \ \ 
    m_i=\left\{
    \begin{aligned}
    1,&\ \bp_{t}\in\mathbf{P} \\
    0,&\ otherwise
    \end{aligned}
    \right.
\end{equation}
where $N_p$ is the total number of unmasked patches, $\mathbf{P}=\left\{ \left( x_t, y_t \right) \mid 0<x_t<W,0<y_t<H \right\}$ represents the points in the video clip.

\section{More Details on Baseline Reconstruction Targets Extraction}
In Section 4.2.2, we introduce different types of baseline reconstruction targets (\eg, HOG, HOF and MBH) and investigate the effect of reconstructing these targets in the mask-and-predict task. In this section, we present the details of extracting these baseline reconstruction targets.

\label{app:other-targets}
\para
\noindent\textbf{HOG}. In our experiment, we utilize the successful application of the HOG feature~\cite{HOG} in masked video modeling~\cite{MaskFeat} to represent appearance information. HOG is the statistic histogram of image gradients in RGB channels. To speed up the calculation, we re-implement the HOG extractor by convolution operator with Sobel kernel and the scatter method in the PyTorch~\cite{PyTorch} Tensor API, and thus the calculation process can be conducted on the GPU. With the exception of the L2 normalization, which does not work in our practice, we precisely follow the recipe in Wei's work~\cite{MaskFeat} to calculate a feature map of the entire image and then fetch the 9-bins HOG feature in each RGB channel.

\para
\noindent\textbf{HOF and MBH}. These two features are spatial-temporal local features that have been widely used in the action recognition task since the 2010s. Similar to the HOG feature, these two features are the histogram of optical flow~\cite{HOF}, the only difference is that MBH is the second-order histogram of the optical flow~\cite{MBH}. We implement them based on the HOG extractor that deals with optical flow.

\section{More Details on Pre-training and Fine-tuning Settings}
We pre-train \sexyname on the two large-scale video datasets (cf. Tab.~\ref{tab:app-pretrain-setting}) and then transfer to four action recognition benchmark datasets (cf. Tab.~\ref{tab:app-finetune-setting}). We linearly scale the base learning rate w.r.t. the overall batch size, $lr=base\_lr\times\frac{batch size}{256}$. More details are presented as follows:

\label{app:settings}
\label{app:dataset}
\para
\noindent\textbf{Something-Something V2.\footnote{\url{https://developer.qualcomm.com/software/ai-datasets/something-something}}} The default settings for pre-training and fine-tuning on Something-Something V2 (SSv2) are shown in Tab.~\ref{tab:app-pretrain-setting} and Tab.~\ref{tab:app-finetune-setting}. 
We take the same recipe as in~\cite{VideoMAE}.

\para
\noindent\textbf{Kinetics-400.\footnote{\url{https://www.deepmind.com/open-source/kinetics}}} The default settings for pre-training and fine-tuning on Kinetics-400 (K400) are shown in Tab.~\ref{tab:app-pretrain-setting} and Tab.~\ref{tab:app-finetune-setting}. We take the same recipe as in~\cite{VideoMAE}.

\para
\noindent\textbf{UCF101.\footnote{\url{https://www.crcv.ucf.edu/data/UCF101.php}}} We pre-train the model on the Kinetics-400 for 400 epochs and then transfer to the UCF101. The default settings of fine-tuning are shown in Tab.~\ref{tab:app-finetune-setting}.

\para
\noindent\textbf{HMDB51.\footnote{\url{https://serre-lab.clps.brown.edu/resource/hmdb-a-large-human-motion-database/}}} The default setting is the same as in UCF101.

\begin{table}[h]
    \centering
    \small
    \begin{tabular}{l|m{2cm}<{\centering}m{2cm}<{\centering}}
         config & SSv2 & K400 \\
         \toprule
         optimizer & \multicolumn{2}{c}{AdamW~\cite{AdamW}} \\
         base learning rate & \multicolumn{2}{c}{1.5e-4} \\
         weight decay & \multicolumn{2}{c}{0.05} \\
         optimizer momentum & \multicolumn{2}{c}{$\beta_1,\beta_2=0.9,0.999$~\cite{Momentum}} \\
         batch size & \multicolumn{2}{c}{768} \\
         learning rate schedule & \multicolumn{2}{c}{cosine decay~\cite{CosineDecay}}\\
         warmup epochs & \multicolumn{2}{c}{40} \\
         flip augmentation & \emph{no} & \emph{yes} \\
         augmentation & \multicolumn{2}{c}{MultiScaleCrop} \\
    \end{tabular}
    \caption{\textbf{Pre-training setting.}}
    \label{tab:app-pretrain-setting}
\end{table}

\begin{table}[h]
    \centering
    \small
    \begin{tabular}{l|m{1cm}<{\centering}m{1cm}<{\centering}m{1cm}<{\centering}}
         config & SSv2 & K400 & Others \\
         \toprule
         optimizer & \multicolumn{3}{c}{AdamW} \\
         base learning rate & \multicolumn{3}{c}{1.5e-4} \\
         weight decay & \multicolumn{3}{c}{0.05} \\
         optimizer momentum & \multicolumn{3}{c}{$\beta_1,\beta_2=0.9,0.999$} \\
         layer-wise lr decay & \multicolumn{3}{c}{0.75~\cite{BEiT}} \\
         batch size & \multicolumn{3}{c}{128}\\
         learning rate schedule & \multicolumn{3}{c}{cosine decay} \\
         warmup epochs & \multicolumn{3}{c}{5} \\
         training epochs & 30 & 75 & 100 \\
         flip augmentation & \emph{no} & \emph{yes} & \emph{yes} \\
         RandAug & \multicolumn{3}{c}{(9,0.5)~\cite{RandAug}} \\
         label smoothing & \multicolumn{3}{c}{0.1~\cite{LabelSmooth}} \\
         mixup & \multicolumn{3}{c}{0.8~\cite{Mixup}} \\
         cutmix & \multicolumn{3}{c}{1.0~\cite{CutMix}} \\
         drop path & \multicolumn{3}{c}{0.1~\cite{DropPath}} \\
         repeated sampling & \multicolumn{3}{c}{2~\cite{ReapSample}}
    \end{tabular}
    \caption{\textbf{Fine-tuning setting.}}
    \label{tab:app-finetune-setting}
\end{table}

\section{Additional Experimental Results on Ablation Studies}
\label{app:add-results}
\para
\noindent\textbf{Training speed-up versus performance decrease.}
To speed up the training procedure during the ablation studies, we (1) randomly select 25\% videos from the training set of Something-Something V2 dataset to pre-train the model, (2) randomly drop 50\% patches following~\cite{MAR}, specifically, we adopt a random masking strategy to select 50\% input patches to reduce the computation of self-attention. We present a comparison across the speed-up and performance decrease due to these techniques in Tab.~\ref{tab:app-sppedup}. Notice that with an acceptable performance drop (-3.7\%), we speed up the ablation experiment by 2.5 times.

\begin{table}[h]
    \centering
    \small
    \begin{tabular}{lcccc}
        \toprule
         Data & Input patches & Acc.@1 & Acc.@5 & Speed-up \\
         \midrule
         100\% & 100\% & \textbf{64.8} & \textbf{89.7} & 1.0$\times$ \\
         25\% & 50\% & 61.1 & 86.9 & \textbf{2.5$\times$} \\
         \bottomrule
    \end{tabular}
    \caption{\textbf{Training speed-up versus performance decrease}. We report the total time cost of the entire pre-training and fine-tuning procedure. We use a random masking strategy to drop a portion of input patches.}
    \label{tab:app-sppedup}
    \vspace{-4mm}
\end{table}

\para
\noindent\textbf{Complexity and necessity of data pre-processing steps}
The data pre-processing includes 4 steps: 1) pre-extracting the optical flow; 2) removing the camera motion; 3) extracting dense trajectories and 4) removing the inaccurate trajectories. 
We only need to pre-process the data one time, taking less than 24 hours, to get simpler and cleaner reconstruction targets. These targets can be reconstructed using a decoder with fewer layers, leading to a faster pretraining process compared with VideoMAE (2.12 vs. 2.24 min/epoch). Besides, removing steps 2 and 4 lowers the performance on SSV2 by $4.4\%$ and $0.7\%$, respectively.

\begin{figure}
    \centering
    \includegraphics[width=0.95\linewidth]{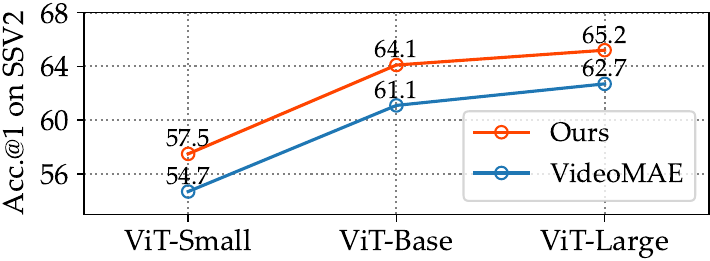}
    \caption{\textbf{Comparing \sexyname and VideoMAE on different ViT backbone.} Our \sexyname scales well on different ViT variants.}
    \label{fig:backbone}
    \vspace{-4mm}
\end{figure}

\para
\noindent\textbf{Scalability on model size of \sexyname.}
We conduct experiments on Something-Something V2 using different ViT variants. In Fig.~\ref{fig:backbone}, both our MME and VideoMAE perform better as the model size grows, and our MME outperforms VideoMAE consistently. This shows the scalability and superiority of our method.

\begin{figure}[h]
\centering
\begin{minipage}[h]{0.485\textwidth}
    \centering
    \includegraphics[width=0.85\linewidth]{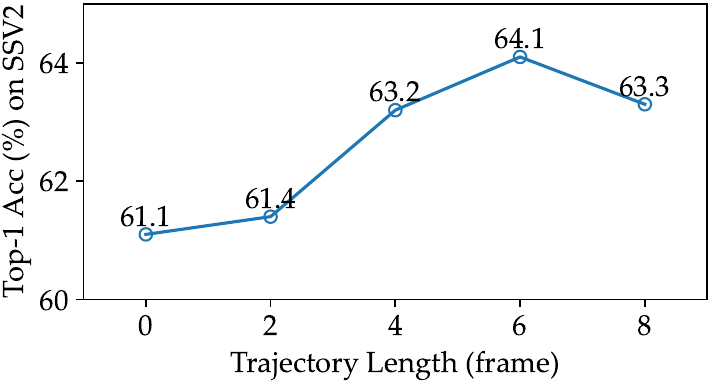}
    \caption{\textbf{Ablation study on trajectory length.}}
    \label{fig:exp-traj-len}
\end{minipage}
\begin{minipage}[h]{0.485\textwidth}
    \centering
    \includegraphics[width=0.85\linewidth]{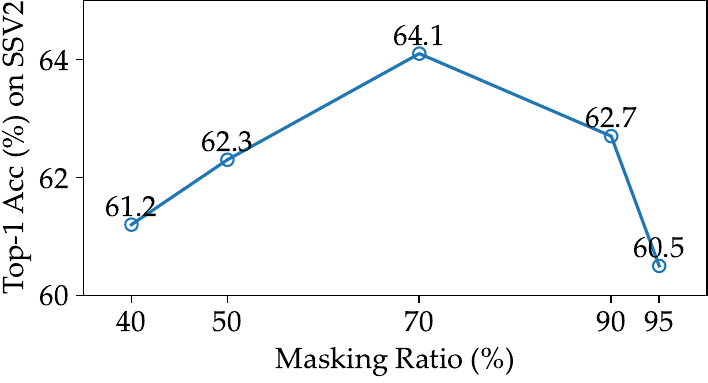}
    \caption{\textbf{Ablation study on masking ratio.}}
    \label{fig:exp-mask-ratio}
\end{minipage}
\end{figure}

\para
\noindent\textbf{Detailed searching results of hyper-parameters.}
We conduct ablation studies on MME design in detail. Specifically, we present detailed hyper-parameters searching results on masking ratio and trajectory length design, as shown in Fig.~\ref{fig:exp-traj-len} and Fig.~\ref{fig:exp-mask-ratio}. Besides, we also complete the ablation results on spatial trajectory density. Extended from Tab.~\ref{tab:traj-density}, we increase the number of trajectories per patch from 4 to 16. Performance on SSV2 drops slightly by 0.7\%, indicating that 4 trajectories are dense enough. 

\para
\noindent\textbf{Comparing with the original version of dense trajectory.}
We introduce the dense trajectories~\cite{dt} into our mask motion modeling task to explicitly represent motion by densely tracking trajectories of different object parts. This video descriptor has been successfully applied in action recognition~\cite{dt}. The iDT~\cite{iDT} further improves the dense trajectories by removing the effects of camera motion, thus enhancing the performance of action recognition.
The original version of iDT combines short-term motion targets, including HOF and MBH features to represent motion. However, we found that only using the position changes to represent objects' movement is enough, see Tab.~\ref{tab:app-combination}.

\begin{table}[h]
    \centering
    \small
    \begin{tabular}{ccc}
        \toprule
         Additional features & Acc.@1 & Acc.@5 \\
         \midrule
         Traj. + MBH + HOF & 63.5 & 88.2 \\
         Traj. (ours) & \textbf{63.5} & \textbf{88.4} \\
         \bottomrule
    \end{tabular}
    \caption{\textbf{Combining short-term motion features in a trajectory volume}. Discarding these additional short-term motion features draws a competitive result.}
    \label{tab:app-combination}
\end{table}

\para
\noindent\textbf{Comparing with the MaskFeat directly.}
We provide a direct comparison with MaskFeat using the MViTv2-S backbone by pre-training on K400 for 300 epochs and then fine-tuning on SSvV2, following the recipe in MaskFeat. In Tab.~\ref{tab:compare-maskfeat}, our MME outperforms the MaskFeat by 1.5\% on SSv2.

\begin{table}[h]
\small
\centering
\makeatletter\def\@captype{table}
    \begin{tabular}{ccc}
        \toprule
        \specialrule{0em}{0.3pt}{0.3pt}
         Method & Backbone & SSv2 Acc.@1 \\
         \midrule
         MaskFeat & MViTv2-S  & 67.7\% \\ 
         Ours & MViTv2-S & \textbf{69.2\% (+1.5\%)} \\ 
         \bottomrule
    \end{tabular}
    \vspace{-2mm}
    \caption{Comparison with MaskFeat on K400 and SSV2.}
    \label{tab:compare-maskfeat}
    \vspace{-4mm}
\end{table}

\section{More Visualization Results}
\label{app:visual}
We present more visualization results of the motion trajectories. Videos are sampled from the Something-Something V2 validation set. For each 3D cube patch shape $2\times16\times16$, we visualize the first frame and the trajectories start from it. All the trajectories are with length $L=6$ as our default setting.

\begin{figure*}
    \centering
    \includegraphics[width=\linewidth]{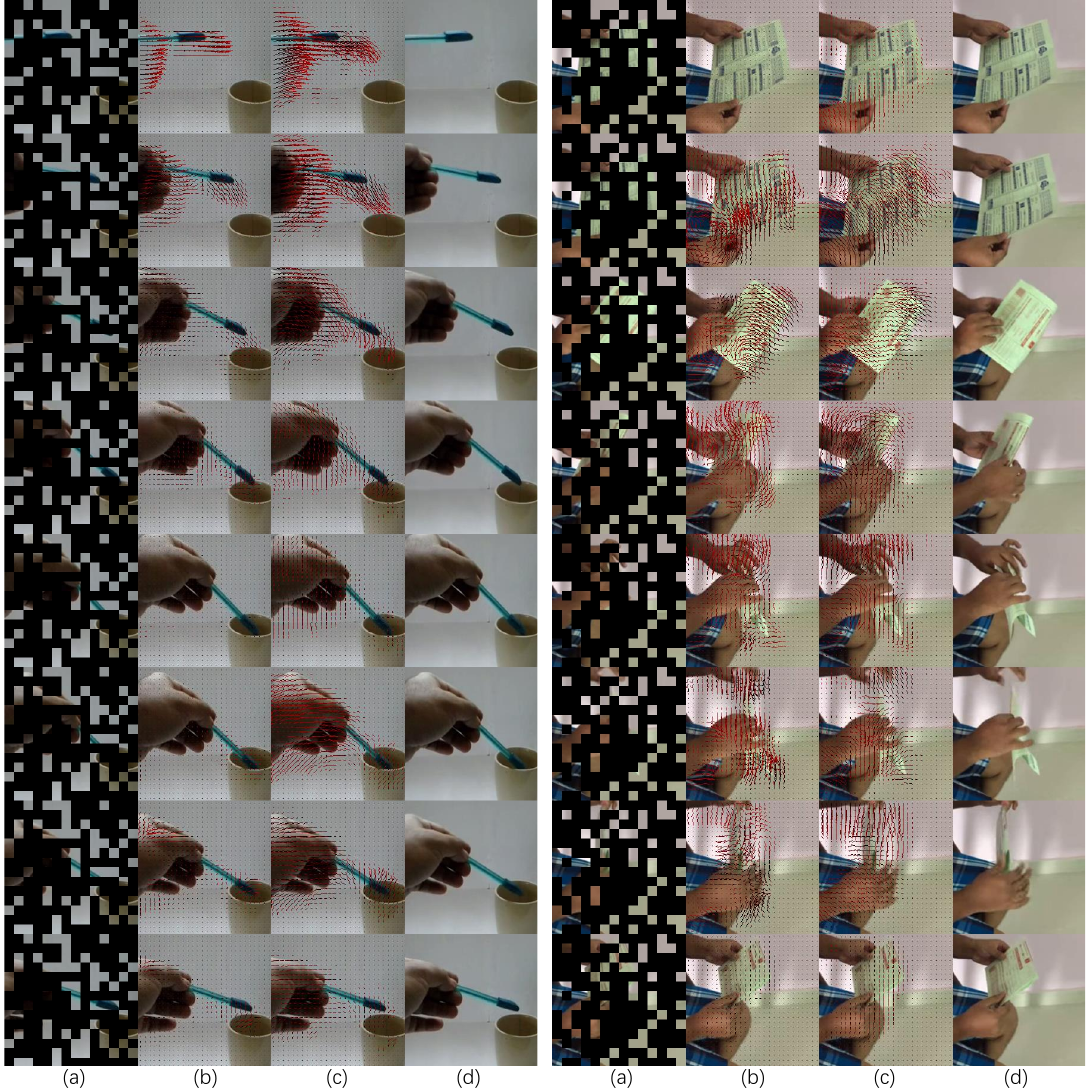}
    \caption{\textbf{Visualization of predicted motion trajectory.} Each column represents: (a) Masked frames; (b) ground truth; (c) prediction; (d) original frame. Trajectories are colored from dark to light in chronological order.}
    \label{fig:extra-1}
\end{figure*}

\begin{figure*}
    \centering
    \includegraphics[width=\linewidth]{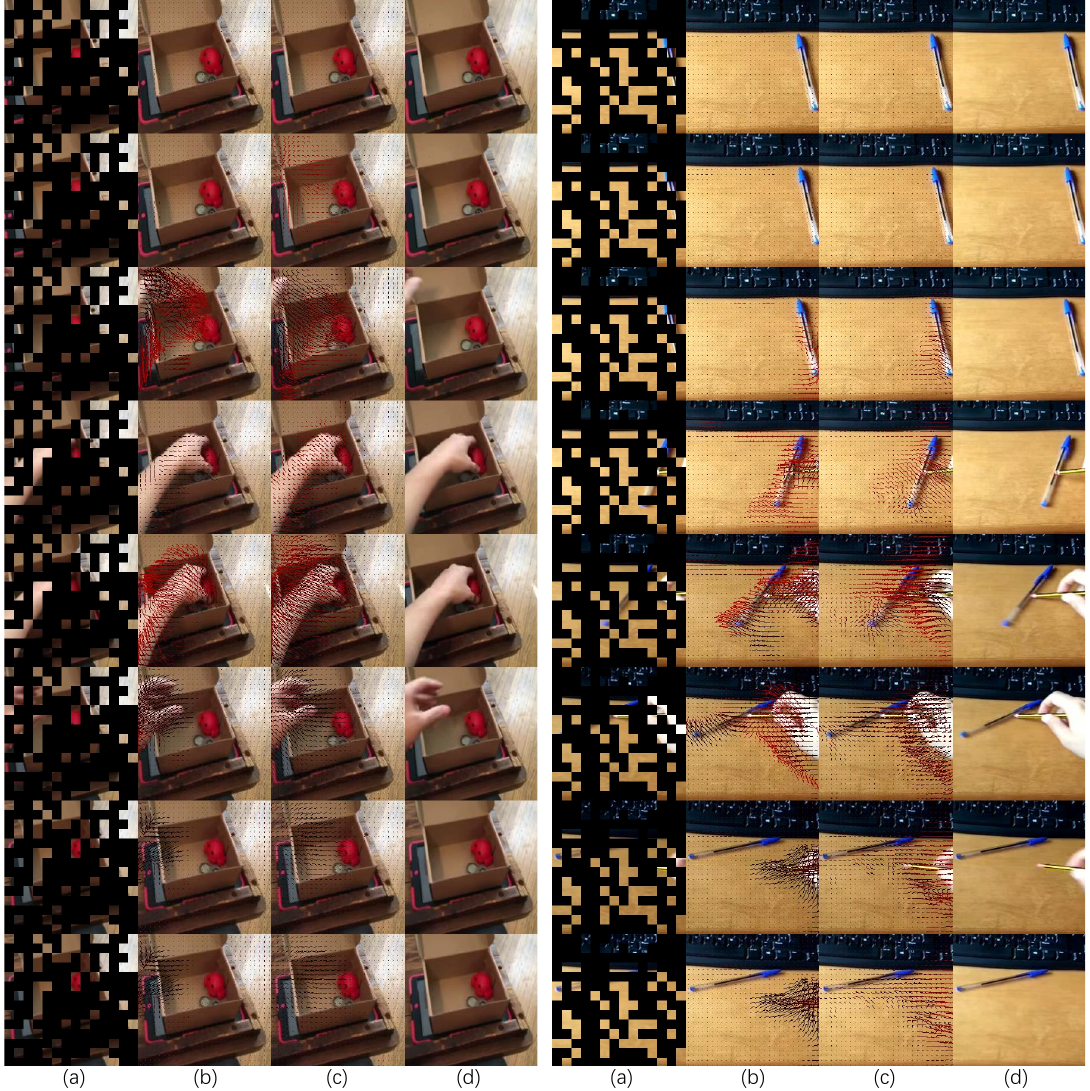}
    \caption{\textbf{Visualization of predicted motion trajectory.} Each column represents: (a) Masked frames; (b) ground truth; (c) prediction; (d) original frame. Trajectories are colored from dark to light in chronological order.}
    \label{fig:extra-2}
\end{figure*}

\end{document}